\documentclass{article}

\usepackage[english]{babel}

\usepackage[letterpaper,top=2cm,bottom=2cm,left=3cm,right=3cm,marginparwidth=1.75cm]{geometry}

\usepackage{amsmath}
\usepackage{graphicx}
\usepackage[colorlinks=true, allcolors=blue]{hyperref}
\usepackage{microtype}
\usepackage{graphicx}
\usepackage{booktabs}
\usepackage{appendix}
\usepackage{subfigure}
\usepackage{cite}
\usepackage{hyperref}
\usepackage{bm}

\usepackage{amsmath}
\usepackage{amssymb}
\usepackage{mathtools}

\title{Learning Stochastic Dynamical Systems as an Implicit Regularization with Graph Neural Networks}
\author{\bf\normalsize{
Jin Guo$^{1,}$\footnotemark[2],
Ting Gao$^{1,}$\footnotemark[1],
Yufu Lan$^{1,}$\footnotemark[3],
Peng Zhang$^{1,}$\footnotemark[4],
Sikun Yang$^{2,}$\footnotemark[5],
Jinqiao Duan$^{1,3,}$\footnotemark[6]
}\\[10pt]
\footnotesize{$^1$ Center for Mathematical Sciences, Huazhong University of Science and Technology, Wuhan 430074, China} \\
\footnotesize{$^2$ Computer Science, Great Bay University, Dongguan, Guangdong 523000, China}\\
\footnotesize{$^3$ Department of Mathematics and Department of Physics, Great Bay University, Dongguan, Guangdong 523000, China}
}

\date{}

\begin{document}
\maketitle

\begin{abstract}
Stochastic Gumbel graph networks are proposed to learn high-dimensional time series, where the observed dimensions are often spatially correlated. To that end, the observed randomness and spatial-correlations are captured by learning the drift and diffusion terms of the stochastic differential equation with a Gumble matrix embedding, respectively. In particular, this novel framework enables us to investigate the implicit regularization effect of the noise terms in S-GGNs. We provide a theoretical guarantee for the proposed S-GGNs by deriving the difference between the two corresponding loss functions in a small neighborhood of weight. Then, we employ Kuramoto's model to generate data for comparing the spectral density from the Hessian Matrix of the two loss functions. Experimental results on real-world data, demonstrate that S-GGNs exhibit superior convergence, robustness, and generalization, compared with state-of-the-arts.
\end{abstract}

\footnotetext[2]{Email: \texttt{jinguo0805@hust.edu.cn}}
\footnotetext[1]{Email: 
 \texttt{tgao0716@hust.edu.cn}}
\footnotetext[3]{Email: \texttt{m202070073@hust.edu.cn}}
\footnotetext[4]{Email: \texttt{kazusa\_zp@hust.edu.cn}}
\footnotetext[5]{Email: \texttt{sikunyang@gbu.edu.cn}}
\footnotetext[6]{Email: \texttt{duan@gbu.edu.cn}}

\section{Introduction}

Multivariate time series enable us to thoroughly investigate the statistical pattern among the obeserved dimensions, and to make predictions. Over last decades, an amount of works have been dedicated to modeling time series data arising in applications including finance, biology, and etc. Real-world time series data often exhibit a certain amount of correlations or causal relationships between the oberved dimensions. For instance, the prices of many financial derivatives, could be simultaneously influenced by the common market signals and interventions. 
The dynamics of electroencephalogram (EEG) brain signals, implicitly reflect a latent graph structure~\cite{diykh2016eeg}. Hence, it is crucial to exploit the graph structure in modelling multivariate time series, within a dynamical system. 

Graph neural networks (GNN)~\cite{wu2020comprehensive} aim to extract both local and global features, by leveraging available graph structure, using neural networks. 
Canonical GNNs, including graph convolutional networks (GCN) \cite{kipf2016semi} and graph attention networks (GAT) \cite{velickovic2017graph}, have demonstrated their strong capacity in capturing graph-structured data. 
Recently, there has been growing interests in predicting time series with graph neural networks to capture the underlying graph structure. For instance, temporal graph convolutional neural networks (T-GCN) \cite{zhao2019t} can integrate temporal information with the available graph structure. Nonetheless, the graph information, which may characterize the correlations or causal relations between observed dimensions, are either not immediately available, or contain noise.

Stochastic dynamical system, as a mathematical tool \cite{duan2015introduction} to describe physical system evolving over time, becomes more and more popular nowadays to model various real world complex phenomena Stochastic dynamical system~\cite{rakthanmanon2012searching}, has been receiving increasing attention specifically in machine learning domain~\cite{tzen2019neural, chen2018neural}, because of its mathematical rigor and strong modelling flexibility. Besides, it is also a bridge connecting real data with deep learning algorithms \cite{zhang2020forward, thorpe2018deep}. On the one hand, many kinds of neural network has the corresponding continuous version of  differential equations, which helps building some convergence guaranteed neural networks such as \cite{NODE,NDE,he2016deep,YANG2023279}. On the other hand, mathematical analysis from dynamical system point of view could also provide insights on loss landscape and critical points \cite{zhang2021embedding}. Furthermore, investigation on Edge of Stability (EOS) phenomenon are also promising research directions \cite{li2022analyzing, sagun2016eigenvalues}.

Moreover, the Gumbel graph neural network (GGN)~\cite{zhang2019general}, is advanced to recover the graph structure underlying time series, within a dynamical system. 
By effectively leveraging the graph structure, the GGN achieves better accurarcy in predicting high-dimensional time series. 
Nevertheless, the GGN suffer from the problems of limited generalization capabilities and excessive smoothness stemming from the inherent properties of graph neural networks themselves.
To address these concerns, this paper proposes the stochastic 
Gumbel Graph Network (S-GGN) model by introducing a diffusion term. The main contributions of this paper can be summarized as follows: 

\begin{itemize}

    \item  A stochastic Gumbel graph network (S-GGN) model is proposed to improve the model robustness and generalization ability in capturing high-dimensional time series, by introducing the noise term. In particular, we thoroughly study the convergence of the S-GGN model with theoretical analysis (Sec.\ref{sec2}).

\item   A grouped convolution S-GGN structure, is advanced to capture \emph{noisy} graph-structured time series.
Using convolution operations, the model can effectively reconstruct the dynamics by leveraging the external node features to remove the noise effects.

\item Experiments on real-world problems~\cite{kuramoto1975self}, demonstrate the superior generalization capability of the S-GGN model without compromising its accuracy, compared with GCNs (Sec.\ref{sec3}). 

\end{itemize}

\section{The S-GGN Frameworks}\label{sec2}

\subsection{S-GGN model}
The GGN, consisting of a network generator and a dynamics learner, recovers the underlying dynamical systems from observations such as high-dimensional time series data. The network generator within GGN utilizes the reparameterization technique known as the Gumbel-softmax trick \cite{jang2016categorical}, whereby the graph is sampled based on probabilities. The application of the Gumbel-softmax trick allows the GGN to directly apply the backpropagation algorithm to calculate the gradient and optimize the network. Based on the connection between the GGN model and the discrete representation of the dynamical system, we extend its applicability to a formulation that aligns with the discrete form of the stochastic dynamical system, called Stochastic Gumbel Graph Neural Network (S-GGN). Thus, the dynamic learner of S-GGN can be represented as
\begin{equation}
    X_{predict}^{t+1}= X^t + f(X^t,A)\Delta t +  \sigma(X^t,A) \xi_t\sqrt{\Delta t}\label{SGNN express}
\end{equation}
where $X^t$ denotes the state vector of all $N$ nodes at moment $t$ and $A$ is the symmetric adjacency matrix constructed by the network generator. Here $\xi_t\sim \mathcal{N}(0,I)$ is an independent standard normal random vector.

The graph neural network module within S-GGN can be depicted as a composition of the following mappings:
\begin{equation}
    \begin{aligned}
        &H_{e_1}^t=f_{v\rightarrow e}(X^t\otimes (X^t)^T),\\
        &H_{e_2}^t=f_e(H_{e_1}^t),\\
        &H_{v_1}^{t+1}=f_{e\rightarrow v}(A*H_{e_2}^t),\\
        &H_{v_2}^{t+1}=f_v(H_{v_1}^{t+1}),\label{four}
    \end{aligned}
\end{equation}
where $X^t\in \mathbf{R}^{N\times d_x}$ denotes the features of $N$ nodes with each node has feature dimension $d$. Let $A$ denotes the adjacency matrix that describes the relationships between the nodes. The $f_{v\rightarrow e},f_e,f_{e\rightarrow v},f_v$ consist of a linear layer and an activation function. The operation $\otimes$ signifies pairwise concatenation, and the symbol $*$ denotes multiplication by elements. The composition of these four mappings in (\ref{four}) corresponds to the function $f$, the same networks' structures as $\sigma$ in equation (\ref{SGNN express}).

Next, we train the two neural networks for functions $f$ and $\sigma$ in (\ref{SGNN express}), denoted as $f_{NN}$ and $\sigma_{NN}$ respectively. Their networks structures are the same with different values.

\subsection{Spectral Analysis}
We denote the parameters of GGN networks as $\bm{\omega}_t$ at the $t$-th interaction. Consider the division $0:=t_0<t_1<\cdots<t_M:=T$ of $[0,T]$, we define that $\delta_m:=t_{m+1}-t_m$. The discretization of parameters' evolution in GGN network can be expressed as
\begin{equation*} 
\bm{\omega}_{t+1}=\bm{\omega}_{t}+f_{NN}(\bm{\omega}_t,X^t)\delta_t,
\end{equation*}
where $f_{NN}:\mathbb{R}^{d_{\bm{\omega}}}\times \mathbb{R}^{d_x}$. After introducing noise $\varepsilon$ to our networks, we consider a rescaling of the noise $\sigma_{NN}\mapsto \varepsilon\sigma_{NN}$, then the following discretization stochastic differential equation (SDE) holds
\begin{equation}    \bm{\omega}_{t+1}^\varepsilon=\bm{\omega}_{t}^\varepsilon+f_{NN}(\bm{\omega}_t^\varepsilon,X_t)\delta_t+\varepsilon\sigma_{NN}(\bm{\omega}_t^\varepsilon,X^t)\xi_t\sqrt{\delta_t}\,,\label{sde}
\end{equation}
where $\sigma_{NN}:\mathbb{R}^{d_{\bm{\omega}}}\times \mathbb{R}^{d_x\times r}$, and $\xi_t\sim \mathcal{N}(0,I)$ is an independent $r$-dimensional standard normal random vector. That is the evolution of parameters in our S-GGN network. Besides, we have the $\bm{\omega}_0^\varepsilon=\bm{\omega}_0$.

Then we give some conditions on drift and diffusion term. To enhance readability, we denote the network functions $f_{NN}$ and $\sigma_{NN}$ as $f$ and $\sigma$ respectively. 

\newtheorem{assumption}{Assumption}[section]
\begin{assumption}\label{assump}
    We assume that the drift term $f$ and diffusion term $\sigma$ satisfy 

    (i) For all $t\in [0,T]$ and $X\in\mathbb{R}^{d_x}$, the maps $\bm{\omega}\mapsto f(\bm{\omega},X)$ and $\bm{\omega}\mapsto \sigma(\bm{\omega},X)$ have Lipschitz continuous partial derivatives in each coordinate up to order three (inclusive).

    (ii) For any $\bm{\omega}\in\mathbb{R}^{d_\omega},t\mapsto f(\bm{\omega},X)$ and $t\mapsto \sigma(\bm{\omega},X)$ are bounded and Borel measurable on $[0, T]$.
\end{assumption}

Under the above assumption, we define the distinct between the loss of S-GGN and GGN networks as
\begin{equation}
    \bm{\mathcal{D}(\bm{\omega})}:= \mathbb{E}[l_{S-GGN}(\bm{\omega}_M^\varepsilon)-l_{GGN}(\bm{\omega}_M)],
\end{equation}
where $l$ denotes the loss function.

\newtheorem{proposition1}[assumption]{Proposition}
\begin{proposition1}\label{thoerem}
(Comparison of the noise induced loss and the deterministic loss) Under Assumption \ref{assump}, the following holds
\begin{equation}
\bm{\mathcal{D}(\bm{\omega})}=\frac{\varepsilon^2}{2}[\hat{R}(\bm{\omega})-\hat{S}(\bm{\omega})]+\mathcal{O}(\varepsilon^3),
\end{equation}
as $\varepsilon\rightarrow 0$, where the $\hat{R}$ and $\hat{S}$ represent
\begin{equation*}
    \begin{aligned}
        &\hat{R}(\bm{\omega})=(\nabla l(\bm{\omega}_M))^T\sum_{k=1}^{M}\delta_{k-1}\hat{\Phi}_{M-1,k}\sum_{m=1}^M\delta_{m-1}\bm{v}_m,\\
        &\hat{S}(\bm{\omega})=\sum_{m=1}^M\delta_{m-1}\mbox{tr}(\sigma_{m-1}^T\hat{\Phi}_{M-1,m}^TH_{\bm{\omega}_M}l\hat{\Phi}_{M-1,k}\sigma_{m-1}),
    \end{aligned}
\end{equation*}
with $\hat{\Phi}_{m,k}:=\hat{J}_m\hat{J}_{m-1}\cdots \hat{J}_k$, the state-to-state Jacobians $\hat{J}_m=I+\delta_m\frac{\partial f}{\partial \bm{\omega}}(\bm{\omega}_m,X_m)$ and the $\bm{v}_m$ is a vector with the p-th component ($p=1,\cdots,d_{\bm{\omega}}$):
$$[\bm{v}_m]^p=\mbox{tr}(\sigma^T_{m-1}\hat{\Phi}_{M-2,m}^TH_{\bm{\omega}}[f_M]^p\hat{\Phi}_{M-2,m}\sigma_{m-1}).$$
Moreover, we have
\begin{equation}
    |\hat{R}(\bm{\omega})|\leq C_R \Delta^2,|\hat{S}(\bm{\omega})|\leq C_S \Delta \label{bound of R and S}
\end{equation}
for $C_R,C_S>0$ independent of $\Delta$, where $\Delta:=\mbox{max}_{m\in \{0,1,\cdots,M-1\}}\delta_m$.
\end{proposition1}
\textbf{Proof.}
    We refer a proof similar to \cite{lim2021noisy}. First, we can apply a Taylor expansion to $\omega_t$, the drift and the diffusion coefficients at a small neighbourhood of $\omega_0$. With Ito formula and comparing the corresponding terms of $\varepsilon$ in the two sides of equation (\ref{sde}), we can obtain the result in \ref{sde}. Next, in conjunction with Lemma 1, 2 and 3 in the literature \cite{lim2021noisy}, it can be demonstrated that (\ref{bound of R and S}) holds.    
  $\,\,\,\,\,\blacksquare$

\newtheorem{remark1}[assumption]{Remark}
\begin{remark1}
    Proposition \ref{thoerem} indicates that introducing noise to the state of a deterministic graph can be considered, on average, as an approximation of a regularized objective functional.
\end{remark1}

\section{Experiments}\label{sec3}
We conduct two experiments to verify the performance of our S-GGN.

\subsection{Kuramoto Model}

The Kuramoto model is a nonlinear model describing the interaction and synchronization of oscillator groups:
\begin{align*}
    \frac{d\theta_i}{dt} = \omega_i + K\sum\limits_{j\neq i}A_{ij}sin(\theta_j - \theta_i), i = 1,2,...,N,
\end{align*}
where $\theta_i$ denotes the $i$-th oscillator phase, $\omega_i$ denotes its natural frequency, $N$ denotes the number of vibrator, $K$ denotes coupling strength which measures the strength of the interaction between the oscillator. Here $A_{ij}\in \{0,1\}$ are the elements of $N \times N$ adjacency matrix. The model takes into account the phase differences of the oscillators and the interactions between them to explain the synchronization phenomenon.

    \begin{figure}[htb]
  \centering
  \includegraphics[width=0.4\columnwidth]{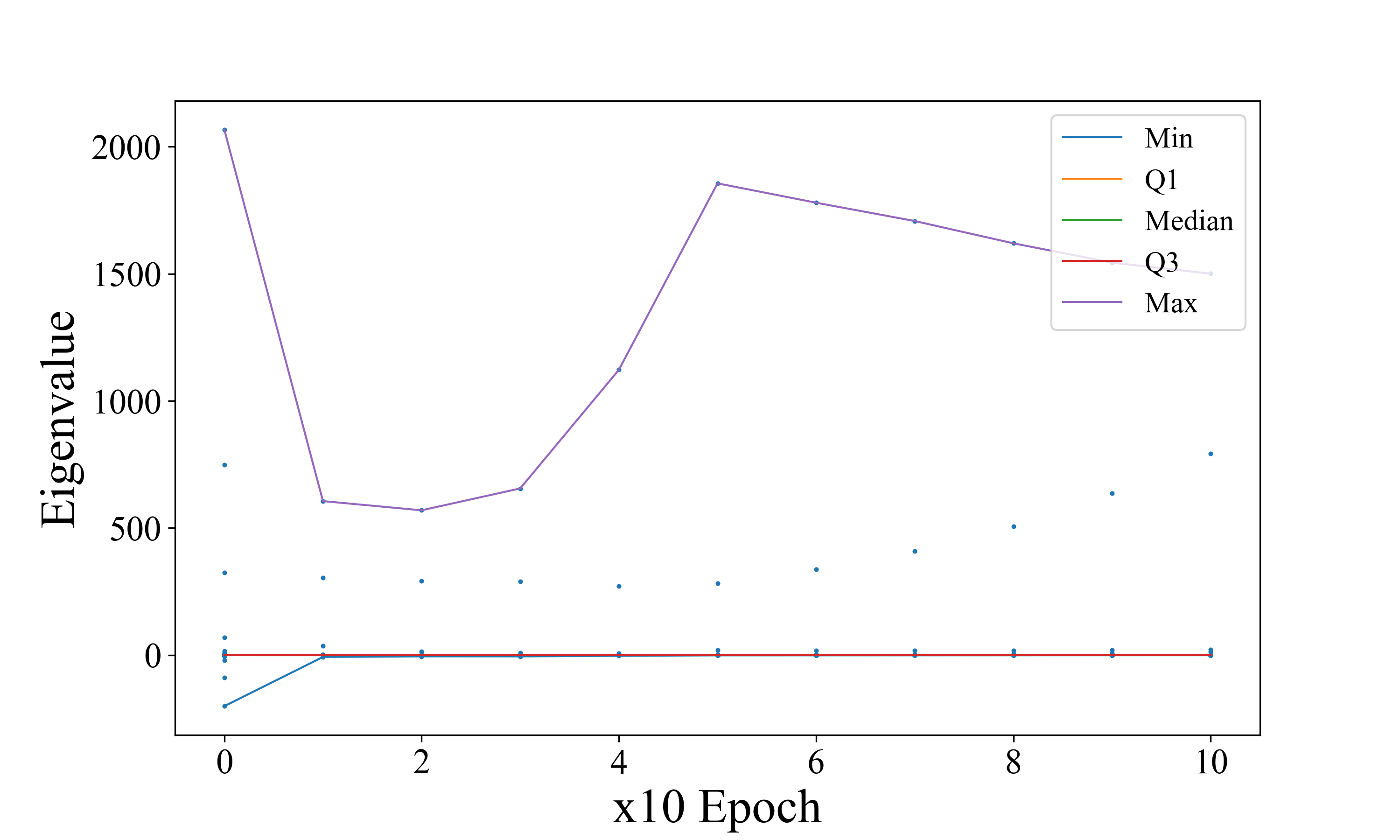} 
  \includegraphics[width=0.4\columnwidth]{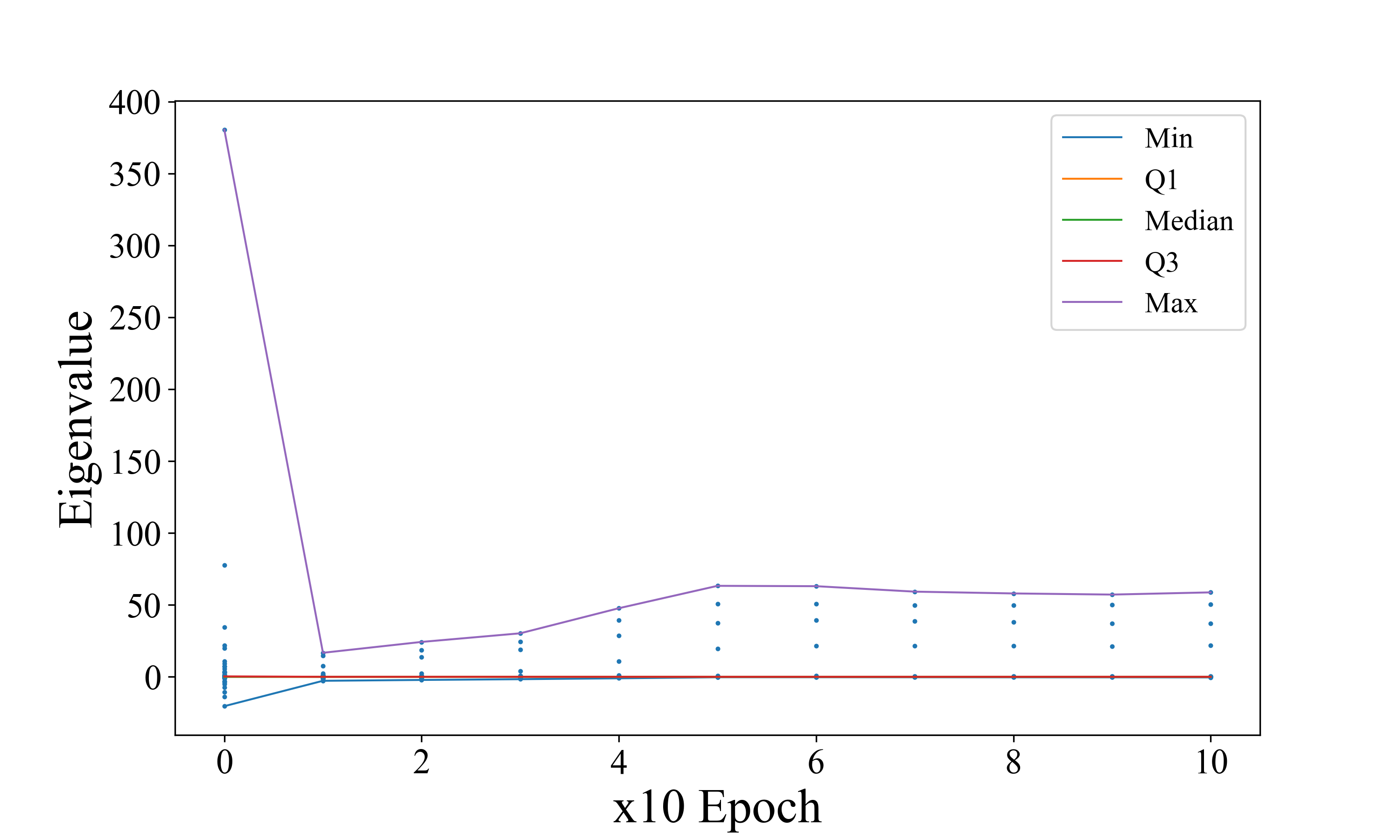}
  \caption{Eigenvalues of the Hessian matrix for GGN (left) and S-GGN (right).}
  \label{Hessian}
\end{figure}

  \begin{figure}[htb]
  \centering
  \includegraphics[width=0.4\columnwidth]{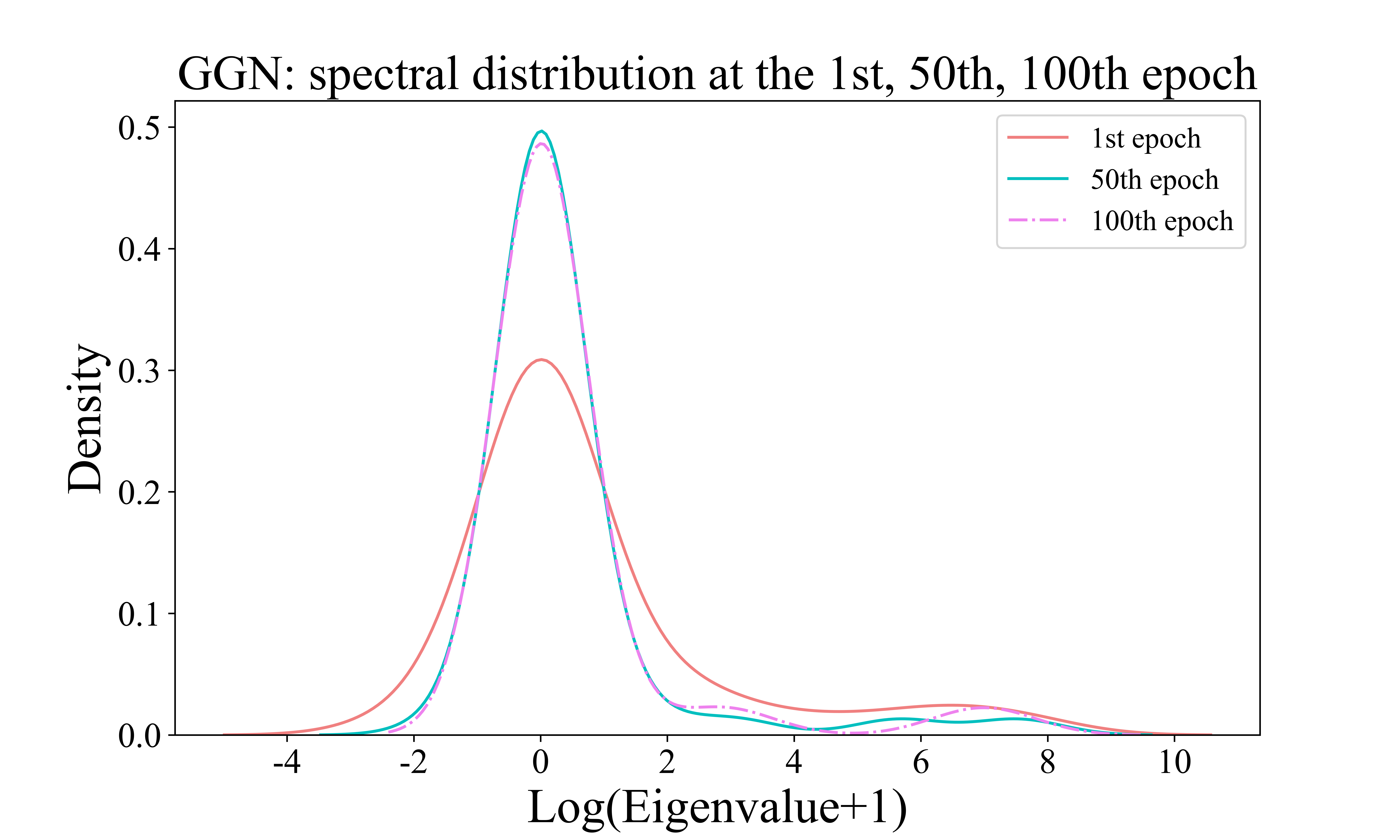} 
  \includegraphics[width=0.4\columnwidth]{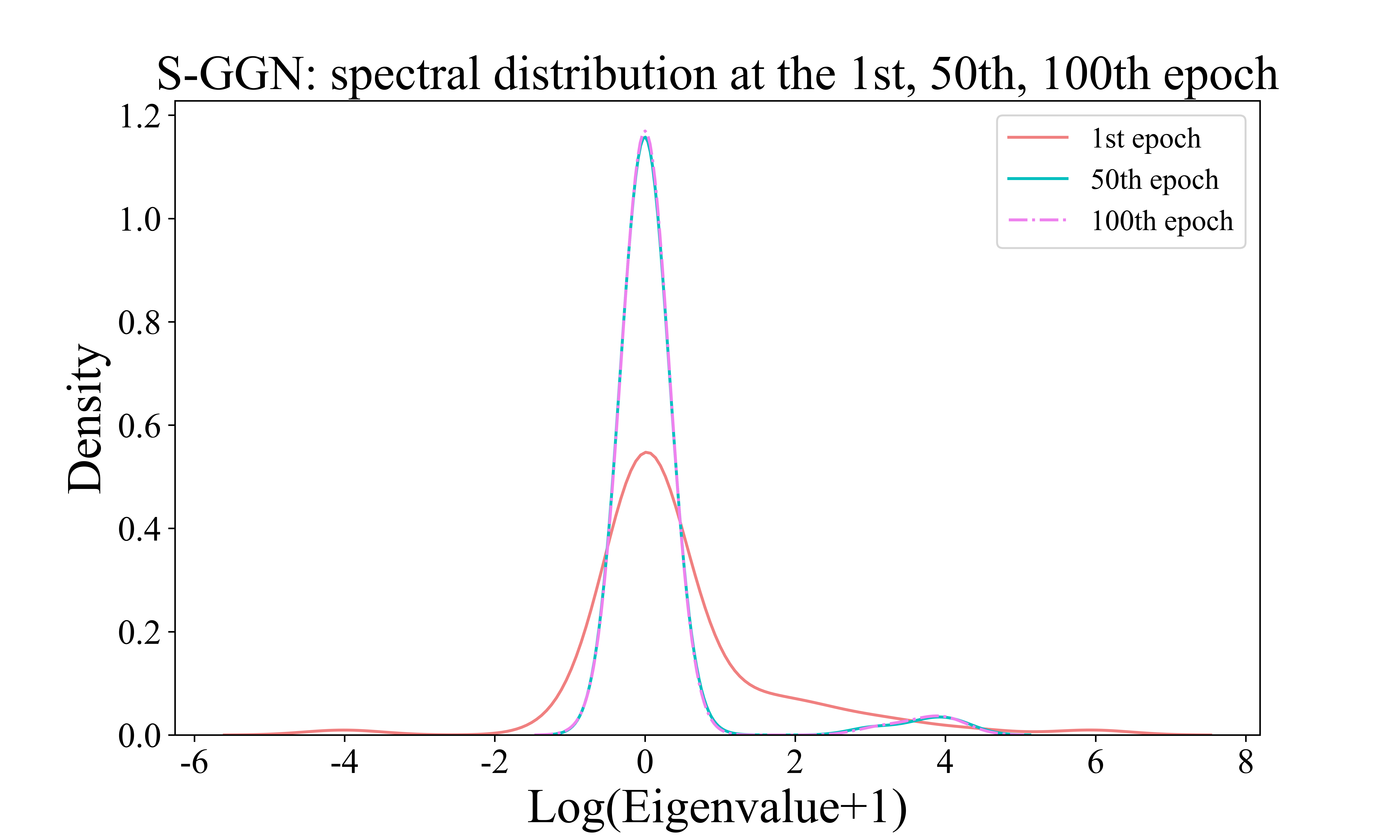}
  \caption{The spectral distribution for GGN (left) and S-GGN (right).}
  \label{spectral}
\end{figure}

Here, we need to initialize the corresponding initial phase and natural frequency for each oscillator, and then calculate the phase of the oscillator attractively according to the above formula.

\begin{itemize}
    \item Data preparation: the numerical solution of the Kuramoto model by the fourth-order Longe-Kutta method.
    \item Data pre-processing: the sin value and frequency of the phase value at the corresponding time, and these two characteristics are taken as the characteristics of the node at the corresponding time. Set the window length to 20.
    \item Experiment settings: the optimizer of network generator and dynamic learner, the number of iteration steps is $3$ and $7$ respectively.
\end{itemize}

For both models, the Hessian Matrix of the empirical loss with respect to weights can be obtained every $10$ epochs, and the corresponding eigenvalues over epochs is shown in the Figure \ref{Hessian}. Observing that S-GGN has to smaller largest eigenvalue, which indicates that S-GGN can find flatter optimal weights, allowing its better performance from sharpness awareness point of view. Figure \ref{spectral} shows the distribution of eigenvalues of the two models after the first, 50-th and 100-th epochs. We can see that the eigenvalues' concentration of S-GGN is much stronger than that of GGN, suggesting different convergence of the two models.

\begin{figure}[h]
    \centering
    \includegraphics[width=0.8\columnwidth]{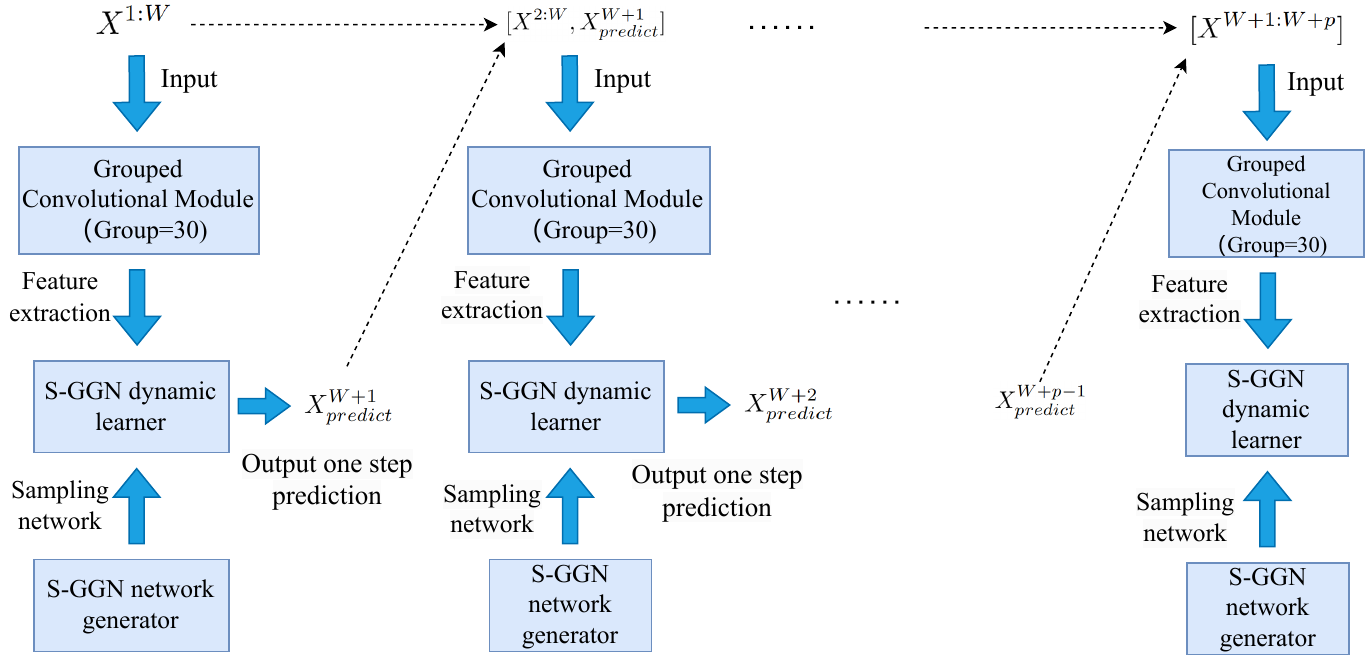}
    \caption{The process of S-GGN Prediction.}
    \label{nstep}
\end{figure}

\subsection{Wireless communication data}

In this experiment, the data is obtained from channel measurements in real-world scenarios. Consider a typical line-of-sight (LOS) scenario with a 24-millisecond spacing between points, and an 8x1 uniform linear array (ULA) at the transmitter end. 

\begin{itemize}
    \item Data standardization: the wireless communication signal data from each base station is decomposed into real and imaginary parts for normalisation.
    \item Data preparation: time window is chosen as $72$ points. Within each time window, features are extracted using group convolution applied to data with a window length of $36$. 
    \item Experimental settings: The Adam optimizer is selected to optimize both the network generator and the dynamic learner, with $3$ and $12$ iteration steps respectively.
\end{itemize}

Due to the highly noisy, non-linear and non-smooth nature of Wireless communication data, direct modelling of the raw signal is challenging.
Convolutional neural networks offer significant advantages in feature extraction, automatically learning local features and retaining spatial structure information. We take a rolling prediction approach wherein a dataset of length $W$ is employed to forecast sequences ranging from $W+1$ to $W+p$. Notably, the actual data ranging from $W+1$ to $W+p$ is not incorporated into the model during this process. The construct is illustrated in Figure \ref{nstep}.

      \begin{figure}[htb]
  \centering
  \includegraphics[width=0.4\columnwidth]{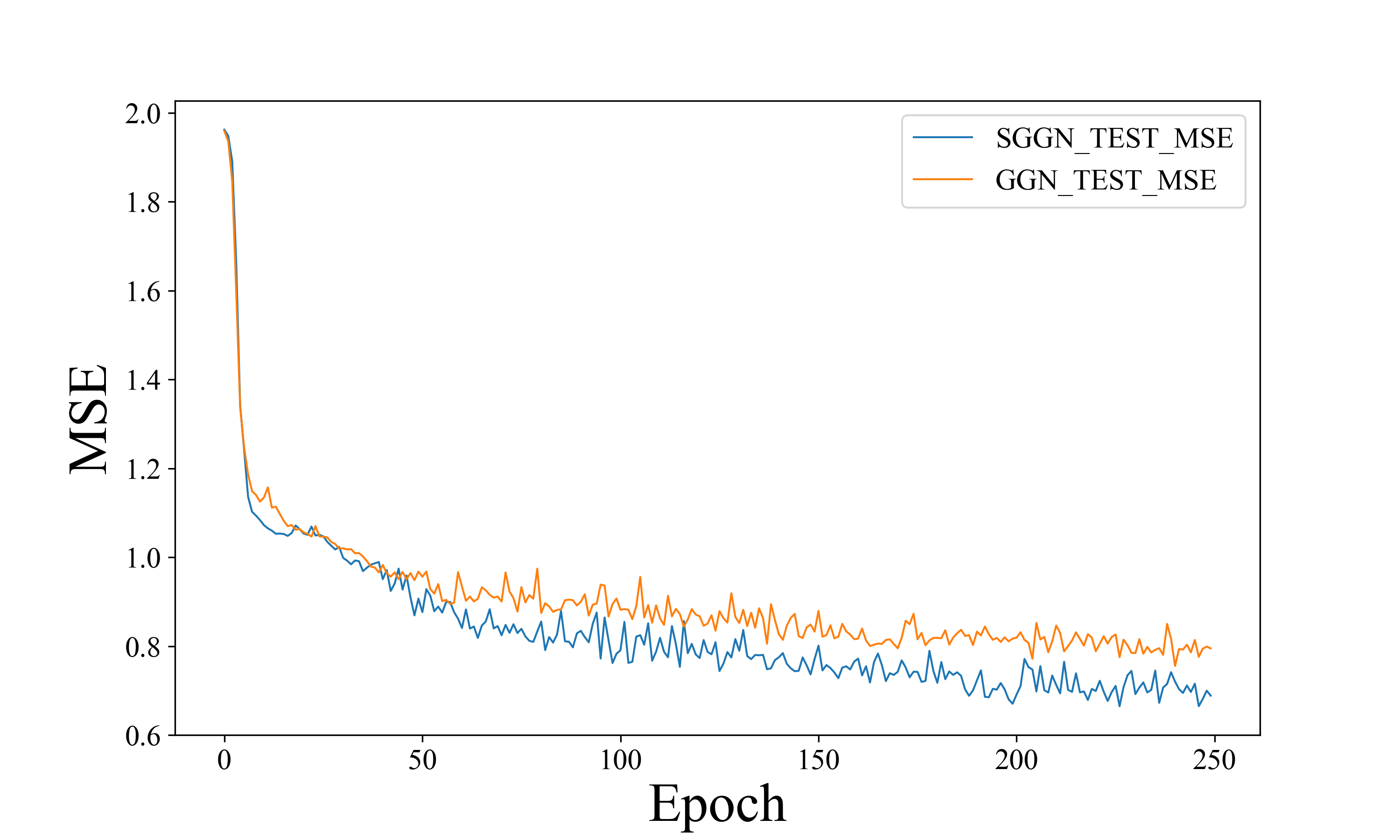} 
  \includegraphics[width=0.4\columnwidth]{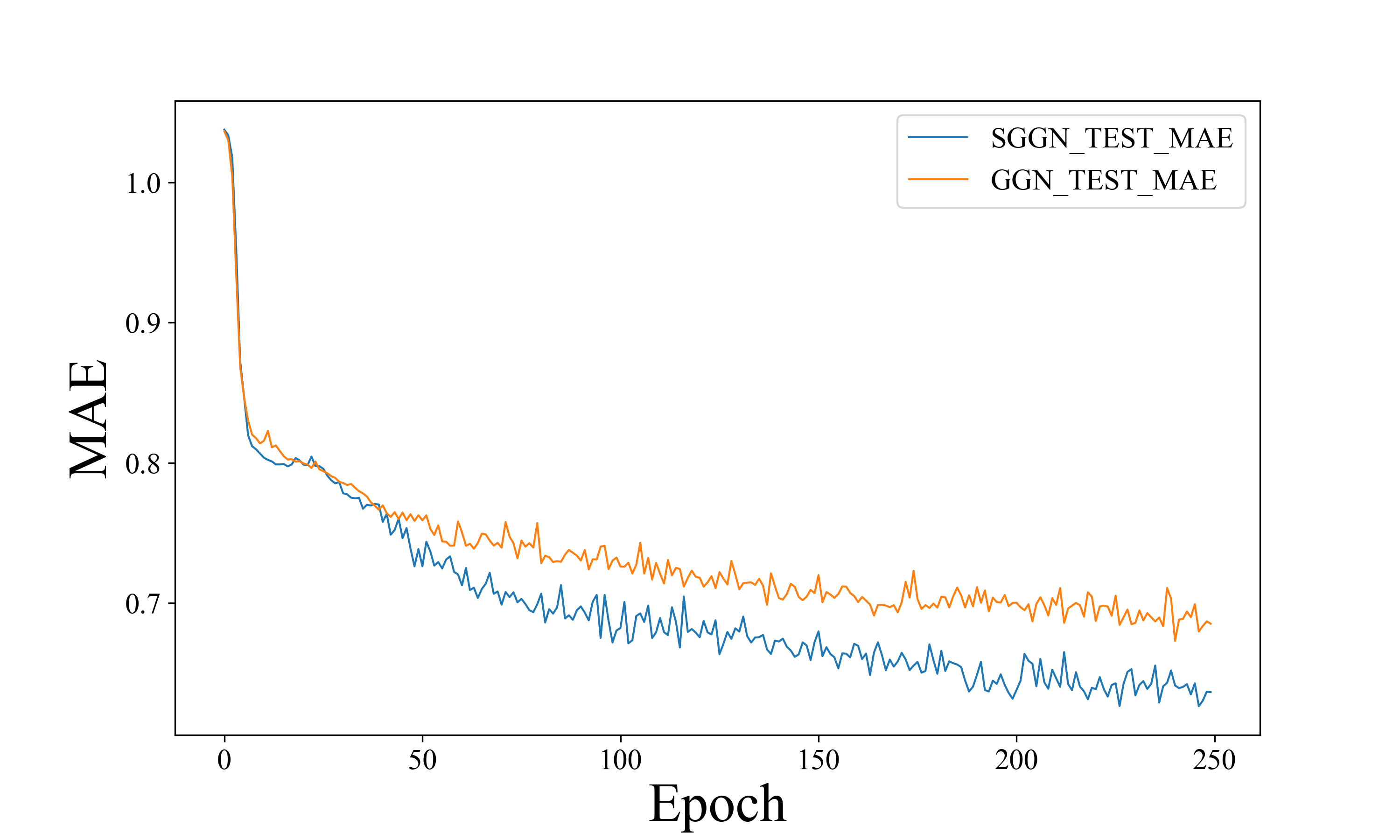}
  \caption{The MSE (left) and MAE (right) for GGN and S-GGN in test set.}
  \label{station}
\end{figure}

The S-GNN model is employed to predict the Wireless communication data.  Figure \ref{station} presents the mean square error and mean absolute error of both the S-GGN and GGN models on the test set. Notably, the S-GGN model demonstrates a smaller error and exhibits superior generalization performance compared to the GGN model.

\begin{figure}[htbp]
    \centering
    \setlength{\abovecaptionskip}{0.cm}  \includegraphics[width=0.8\columnwidth]{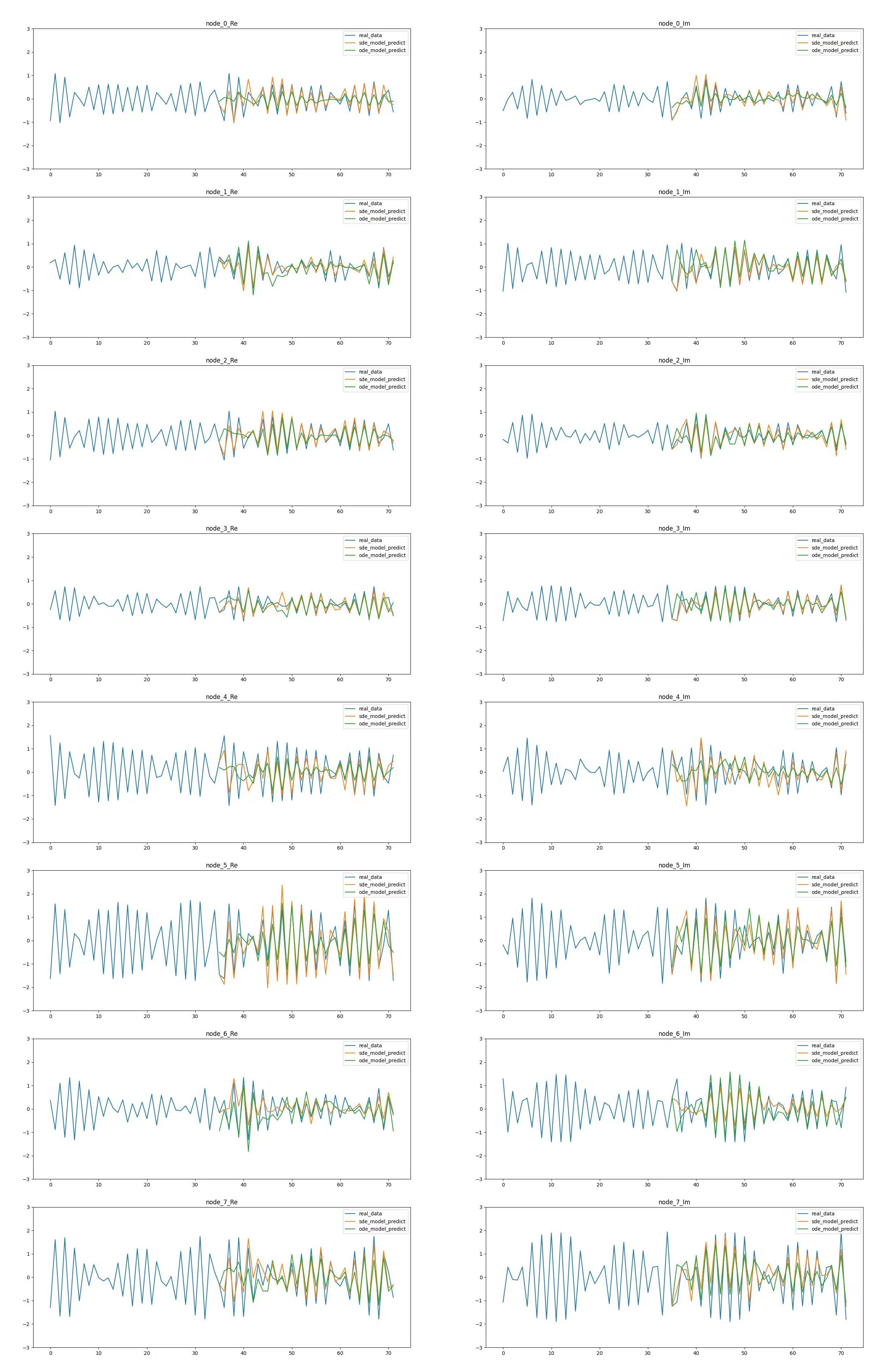}
    \caption{Predictions for 8 base stations signals.}
    \label{8 base stations signals}  
        \vspace{-0.6cm}
\end{figure}

Figure \ref{8 base stations signals} exhibits the comparative prediction outcomes of the GGN and the S-GGN models, utilizing a training sample on the test set. Both the real and imaginary components of the signals originating from the eight node base stations are plotted. The solid blue line represents the actual data, while the solid yellow line corresponds to the prediction results of the S-GGN model, and the solid green line represents the prediction results of the GGN model. We can see the S-GGN outperforms and its prediction result is more close to the true data.

\section{Conclusion}

Considering the complexity of the noise and underlying dynamics of the data, we bring Stochastic Dynamical Systems as a tool to address this problem. First, by comparing the loss of S-GGN and GGN, we can see the term of the multiplicative noise can be treated as a regularization term for the perturbation in a small neighborhood of neural network’s weights. As a result, the S-GGN model could achieve better generalization capabilities and robustness on noisy data. Second, we aim to explore the spectral density over iteration steps for Hessian Matrix eigenvalues of the empirical loss w.r.t. weights. We conduct experiments on data from Kuramoto model to verify the effectiveness of S-GGN. Finally, in real-world applications such as wireless communication data, we introduce group convolution techniques as our data preprocessing, which helps us to get better long-term prediction results. 
Despite this, there are still some problems that need to be solved which we would like to further notice, such as loss analysis through a sharpness awareness point of view, more applications in complex spatial-temporal data like EEG signals in the brain, financial data, molecular dynamics, climate forecasting and so on.

\section*{ACKNOWLEDGMENTS}
This work was supported by the National Key Research and Development Program of China (No. 2021ZD0201300), the National Natural Science Foundation of China (No. 12141107), the Fundamental Research Funds for the Central Universities (5003011053).

\section*{Data availability}
The data that support the findings of this study are
available in GitHub at https://github.com/xiaolangege/sggn.

\bibliographystyle{unsrt}
\bibliography{main}

\begin{thebibliography}{10}

\bibitem{diykh2016eeg}
Mohammed Diykh, Yan Li, and Peng Wen.
\newblock Eeg sleep stages classification based on time domain features and
  structural graph similarity.
\newblock {\em IEEE Transactions on Neural Systems and Rehabilitation
  Engineering}, 24(11):1159--1168, 2016.

\bibitem{wu2020comprehensive}
Zonghan Wu, Shirui Pan, Fengwen Chen, Guodong Long, Chengqi Zhang, and S~Yu
  Philip.
\newblock A comprehensive survey on graph neural networks.
\newblock {\em IEEE transactions on neural networks and learning systems},
  32(1):4--24, 2020.

\bibitem{kipf2016semi}
Thomas~N Kipf and Max Welling.
\newblock Semi-supervised classification with graph convolutional networks.
\newblock {\em arXiv preprint arXiv:1609.02907}, 2016.

\bibitem{velickovic2017graph}
Petar Velickovic, Guillem Cucurull, Arantxa Casanova, Adriana Romero, Pietro
  Lio, Yoshua Bengio, et~al.
\newblock Graph attention networks.
\newblock {\em stat}, 1050(20):10--48550, 2017.

\bibitem{zhao2019t}
Ling Zhao, Yujiao Song, Chao Zhang, Yu~Liu, Pu~Wang, Tao Lin, Min Deng, and
  Haifeng Li.
\newblock T-gcn: A temporal graph convolutional network for traffic prediction.
\newblock {\em IEEE transactions on intelligent transportation systems},
  21(9):3848--3858, 2019.

\bibitem{duan2015introduction}
Jinqiao Duan.
\newblock {\em An introduction to stochastic dynamics}, volume~51.
\newblock Cambridge University Press, 2015.

\bibitem{rakthanmanon2012searching}
Thanawin Rakthanmanon, Bilson Campana, Abdullah Mueen, Gustavo Batista, Brandon
  Westover, Qiang Zhu, Jesin Zakaria, and Eamonn Keogh.
\newblock Searching and mining trillions of time series subsequences under
  dynamic time warping.
\newblock In {\em Proceedings of the 18th ACM SIGKDD international conference
  on Knowledge discovery and data mining}, pages 262--270, 2012.

\bibitem{tzen2019neural}
Belinda Tzen and Maxim Raginsky.
\newblock Neural stochastic differential equations: Deep latent gaussian models
  in the diffusion limit.
\newblock {\em arXiv preprint arXiv:1905.09883}, 2019.

\bibitem{chen2018neural}
Ricky~TQ Chen, Yulia Rubanova, Jesse Bettencourt, and David~K Duvenaud.
\newblock Neural ordinary differential equations.
\newblock {\em Advances in neural information processing systems}, 31, 2018.

\bibitem{zhang2020forward}
Linan Zhang and Hayden Schaeffer.
\newblock Forward stability of resnet and its variants.
\newblock {\em Journal of Mathematical Imaging and Vision}, 62:328--351, 2020.

\bibitem{thorpe2018deep}
Matthew Thorpe and Yves van Gennip.
\newblock Deep limits of residual neural networks.
\newblock {\em arXiv preprint arXiv:1810.11741}, 2018.

\bibitem{NODE}
Ricky T.~Q. Chen, Yulia Rubanova, Jesse Bettencourt, and David~K Duvenaud.
\newblock Neural ordinary differential equations.
\newblock In {\em Advances in Neural Information Processing Systems},
  volume~31, 2018.

\bibitem{NDE}
Patrick Kidger.
\newblock On neural differential equations, 2022.

\bibitem{he2016deep}
Kaiming He, Xiangyu Zhang, Shaoqing Ren, and Jian Sun.
\newblock Deep residual learning for image recognition.
\newblock In {\em Proceedings of the IEEE conference on computer vision and
  pattern recognition}, pages 770--778, 2016.

\bibitem{YANG2023279}
Luxuan Yang, Ting Gao, Yubin Lu, Jinqiao Duan, and Tao Liu.
\newblock Neural network stochastic differential equation models with
  applications to financial data forecasting.
\newblock {\em Applied Mathematical Modelling}, 115:279--299, 2023.

\bibitem{zhang2021embedding}
Y.~Zhan, Z.~Zhang, T.~Luo, and Z.~J. Xu.
\newblock Embedding principle of loss landscape of deep neural networks.
\newblock {\em Advances in Neural Information Processing Systems},
  34:14848--14859, 2021.

\bibitem{li2022analyzing}
Z.~Li, Z.~Wang, and J.~Li.
\newblock Analyzing sharpness along gd trajectory: Progressive sharpening and
  edge of stability.
\newblock {\em arXiv preprint arXiv:2207.12678}, 2022.

\bibitem{sagun2016eigenvalues}
L.~Sagun, L.~Bottou, and L.~LeCun.
\newblock Eigenvalues of the hessian in deep learning: Singularity and beyond.
\newblock {\em arXiv preprint arXiv:1611.07476}, 2016.

\bibitem{zhang2019general}
Zhang Zhang, Yi~Zhao, Jing Liu, Shuo Wang, Ruyi Tao, Ruyue Xin, and Jiang
  Zhang.
\newblock A general deep learning framework for network reconstruction and
  dynamics learning.
\newblock {\em Applied Network Science}, 4(1):1--17, 2019.

\bibitem{kuramoto1975self}
Yoshiki Kuramoto.
\newblock Self-entrainment of a population of coupled non-linear oscillators.
\newblock In {\em International Symposium on Mathematical Problems in
  Theoretical Physics: January 23--29, 1975, Kyoto University, Kyoto/Japan},
  pages 420--422. Springer, 1975.

\bibitem{jang2016categorical}
Eric Jang, Shixiang Gu, and Ben Poole.
\newblock Categorical reparameterization with gumbel-softmax.
\newblock {\em arXiv preprint arXiv:1611.01144}, 2016.

\bibitem{lim2021noisy}
Soon~Hoe Lim, N~Benjamin Erichson, Liam Hodgkinson, and Michael~W Mahoney.
\newblock Noisy recurrent neural networks.
\newblock {\em Advances in Neural Information Processing Systems},
  34:5124--5137, 2021.

\end{thebibliography}

\end{document}